\pdfoutput=1
\documentclass{article}

\usepackage{microtype}
\usepackage{subfigure}
\usepackage{booktabs}
\usepackage{bm}

\usepackage{hyperref}

\usepackage[accepted]{icml2018}

\usepackage{mathtools}
\mathtoolsset{showonlyrefs=true}
\usepackage{amsfonts}

\DeclareMathOperator*{\siblings}{siblings}
\DeclareMathOperator*{\parent}{parent}
\DeclareMathOperator*{\roooot}{root}

\DeclareMathOperator*{\argmax}{argmax}

\DeclareMathOperator*{\relu}{ReLU}
\DeclareMathOperator*{\pseudohuber}{PH}
\newcommand{\eat}[1]{}

\newcommand\etc{\textit{etc.}}
\newcommand\eg{\textit{e.g.}}
\newcommand\ie{\textit{i.e.}}

\newcommand\lcb{\left\{}
\newcommand\rcb{\right\}}
\newcommand\lnb{\left(}
\newcommand\rnb{\right)}

\newcommand\abs[1]{\left|#1\right|}

\newcommand{\thetitle}{Neural Dynamic Programming for Musical Self Similarity}

\newtheorem{definition}{Definition}

\newcommand\Ocal{\mathcal{O}}

\newcommand{\ds}{D_{\text{s}}}

\newcommand{\fembedding}{\ensuremath{f_{E}}}
\newcommand{\fedit}{\ensuremath{f_{SD}}}
\newcommand{\fdeletion}{\ensuremath{f_{D}}}
\newcommand{\gdeletion}{\ensuremath{g_{D}}}
\newcommand{\fsubstitution}{\ensuremath{f_{S}}}
\newcommand{\gsubstitution}{\ensuremath{g_{S}}}
\newcommand{\faddition}{\ensuremath{f_{A}}}
\newcommand{\fscoring}{\ensuremath{f_{W}}}
\newcommand{\fscore}{\fscoring}
\newcommand{\fscorer}{\fscoring}
\newcommand{\fanalogy}{\ensuremath{f_{G}}}
\newcommand{\ganalogy}{\ensuremath{g_{G}}}
\newcommand{\fforecasting}{\ensuremath{f_{F}}}

\newcommand{\realset}{\mathbb{R}}
\newcommand{\dinit}{D_0}
\newcommand{\ttran}{^\top}
\newcommand\dmax{\ensuremath{d_{\text{max}}}}
\newcommand\npriority{\ensuremath{n_{\text{priority}}}}
\newcommand\naive{na\"ive}
\newcommand\naively{na\"ively}

\begin{document}

\twocolumn[
\icmltitle{\thetitle}



\icmlsetsymbol{equal}{*}

\begin{icmlauthorlist}
\icmlauthor{Christian J. Walder}{csiro,anu}
\icmlauthor{Dongwoo Kim}{anu,dtd}
\end{icmlauthorlist}

\icmlaffiliation{csiro}{CSIRO Data61, Black Mountain, Australia}
\icmlaffiliation{anu}{The Australian National University}
\icmlaffiliation{dtd}{Data to Decisions CRC, Kent Town, SA, Australia}

\icmlcorrespondingauthor{Christian}{christian.walder@data61.csiro.au}

\icmlkeywords{Machine Learning}

\vskip 0.3in
]




\printAffiliationsAndNotice{} 


\begin{abstract}
We present a neural sequence model designed specifically for symbolic music. The model is based on a learned edit distance mechanism which generalises a classic recursion from computer science, leading to a neural dynamic program. Repeated motifs are detected by learning the transformations between them. We represent the arising computational dependencies using a novel data structure, the edit tree; this perspective suggests natural approximations which afford the scaling up of our otherwise cubic time algorithm. We demonstrate our model on real and synthetic data; in all cases it out-performs a strong stacked long short-term memory benchmark.
\end{abstract}

\section{Introduction}

A popular approach to symbolic music modelling is to represent pieces by sequences and model them with a Recurrent Neural Network (RNN) \cite{eckschmidhuber,aimusicsurvey,rlmusic,pachetvariations,pachetsurvey,biaxial}. The RNN architecture is typically based on the Long Short-Term Memory (LSTM) \cite{lstm}.

Such models are useful for algorithmic composition and accompaniment; \eg\ one may conditionally sample with human input constraints \cite{pachetmarkovconstraints2011,pfrnn}. Without such constraints, sampling from an LSTM tends to produce music with an undesirable meandering quality that is unlike human composed music (see \eg\ \cite{rlmusic} for a discussion of the issue). 

\begin{figure}

\begin{center}
\includegraphics[width=0.43\textwidth]{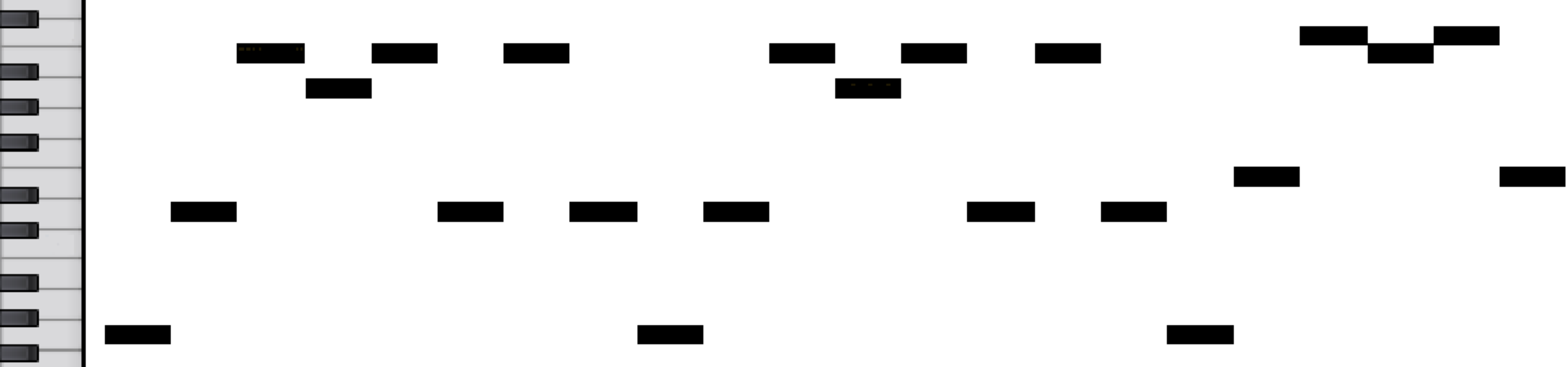}
\end{center}
\vspace{-2mm} 
{\small \hspace{25.6mm}$j$\hspace{48.8mm}{$i$}}
\caption{
\label{fig:bach} 
An excerpt from Bach's \textit{BWV 1007} as a \textit{piano roll} (horizontal $=$ time; vertical $=$ pitch as per the keyboard at left), which ends during the third repetition of a \textit{motif}. By \textit{analogy} with the earlier repeats, the last two notes of the excerpt will likely repeat immediately. Note that the third cycle of the motif introduces a non-trivial \textit{diatonic} (within musical scale) shift of the upper notes. The indices $i$ and $j$ above are referred to in \autoref{sec:model:forecast:analogy}.
}
\end{figure}

Rather, human music is strongly self similar \cite{pareyonbook}; a few motifs may form the building blocks for an entire piece  --- see \eg\ movement one of Beethoven's \textit{Pastoral Symphony} and its entertaining analysis by \citet{bernsteinthree}. These motific building blocks are not simply repeated verbatim throughout a piece of music; rather, as in the example of \autoref{fig:bach}, the fragments are transformed in a musically logical way. This logic demands coherent harmony, melody and phrasing structure; this in turns requires such non-trivial motif transformations as transposition, diatonic shifting, deletion, insertion of passing tones, truncation, \etc

The logic of motif transformation is our focus. Like any sequence model, our model may be factorised causally as in \eqref{eqn:chainrule} of \autoref{sec:model} below. The key idea is that whenever, as we are generating a sequence, we detect earlier point(s) which are motifically related, then we may expect the future to unfold analogously. This is similar in spirit to the concept of \textit{attention} \cite{attention}, which has spurred lots of research interest, especially in sequence to sequence modelling for machine translation. Attentional sequence models are far less common however --- the single example we are aware of is \cite{autoattention}. Indeed, \naively\ modelling sequences with attention suffers from a quadratic time complexity, and is arguably less appropriate for language modelling than for machine translation.

Our model explicitly compares all sub-sequences of all lengths within the very sequence  being modelled, using a dynamic program (DP) that aligns by explicitly modelling generalised edit operations; a \naive\ implementation has a cubic time complexity. This is a challenging example of recent research investigating structured attention and neural (or differentiable) dynamic programming; \eg\ \cite{palma,neuralner,structuredattention,softdtw,ddp}. The computational structure of our algorithm may however be represented with a novel data structure which we call an \textit{edit tree}. By only partially expanding the edit tree, we scale up sufficiently to out-perform the LSTM on synthetic and real music data. 

The structure of the paper is as follows. We complete this introduction by introducing in \autoref{sec:notation}
 the notation we use for both sequences and neural networks. Section \ref{sec:ed} introduces the notion of edit distance, along with a well known DP for computing it. In \autoref{sec:ss} we present a generic algorithm for sequence modelling, which involves comparing a sequence to itself using the edit distance. Section \ref{sec:model} introduces our MotifNet algorithm, which generalises all aspects of the previous section using appropriate neural network sub-modules. We introduce the edit tree in \autoref{sec:trees}, and propose an efficient MotifNet approximation based on it. Section \ref{sec:experiments} provides experiments, and \autoref{sec:conclusion} summarizes our findings and contribution.

\subsection{Notation}
\label{sec:notation}

\textbf{Sequences.} We consider throughout ordered sequences $T= t_1 \cdot t_2 \,\cdots\, t_{\abs{T}}$ of elements $t_i\in\Sigma$, where dots denote concatenation. The subsequence $t_i \,\cdots\, t_j, j\geq i$ of $T$ is denoted $T(i:j)$, and has length $\abs{T(i:j)}=j-i+1$. Similarly $T(:j)$ is the length $j$ prefix of $T$. We let our subsequence notation implicitly handle indices which are out of range by assuming that $T(i:j)=T(\max(1,i):\min(\abs{T},j))$. The empty sequence is represented by $\epsilon$. We abuse the notation by treating length one sequences like elements and vice versa, where appropriate.

\textbf{Neural networks.} Denote by $L(A)$ the linear transformation parametrised by $A\in\realset^{a\times b}$ defined by $L(A): \realset^a \rightarrow \realset^b; \bm{x}\mapsto L(A)(\bm{x}) = A\bm{x}$. Denote the leaky rectified linear unit by $\relu_{\alpha}(x)=\max(x,\alpha x)$ which we abuse the notation by allowing to operate element-wise on vectors. Denote the softmax by $\sigma$ so that $\sigma(\bm{z})_i = \exp(z_i) / (\sum_j \exp(z_j))$. We may now notate \eg\ a two layer feed-forward neural network classifier using the composition operator, \eg\ $\sigma \circ\, L(A_2) \circ \relu_{\alpha} \circ\, L(A_1)$. To obtain symmetry, we employ the \textit{pseudo-Huber}  $\pseudohuber(x,\delta)=\delta^2\lnb\sqrt{1+(x/\delta)^2}-1\rnb$, a smoothed absolute value, in \autoref{sec:model:distance}
. 

\section{Edit Distance}
\label{sec:ed}
Our model features a learned generalization of the notion of edit distance between strings.
The edit distance is a widely used dis-similarity measure between sequences.

\begin{definition}[edit operations]
Operations $\epsilon\rightarrow a$, $a\rightarrow\epsilon$ and $a \rightarrow b$, for $a, b \in \Sigma, a\neq b$ are called \textbf{insertion}, \textbf{deletion} and \textbf{substitution} operations, respectively. Operation $a\rightarrow a$ is called the \textbf{identity} operation.
\end{definition}

\begin{definition}[trace]
A trace of $A$ and $B$ is any sequence $\Gamma=x_1\rightarrow y_1\cdot x_2\rightarrow y_2 \, \cdots \, x_h\rightarrow y_h$ of edit operations such that $A=x_1 \cdot x_2 \,\cdots\, x_h$ and $B=y_1 \cdot y_2 \,\cdots\, y_h$. Note that $\epsilon$ allows insertion/deletion and $\abs A\neq\abs B$ in general.
\end{definition}

\begin{definition}[edit distance]
Let $c(\gamma_i)$ be the non-negative cost associated with edit operation $\gamma$, and let $C(\Gamma) = \sum_{1 \leq i \leq \abs{\Gamma}} c(\gamma_i)$ be the cost of a trace. The edit distance $d(A, B)$ is the minimum cost of a trace of $A$ and $B$.
\end{definition}

\subsection{Dynamic Program for Edit Distance}
\label{sec:ed:wf}

Given two sequences $P$ and $T$, let $D(i,j)$ be the minimum edit distance between prefixes $P(:i)$ and $T(:j)$, for $0\leq i \leq \abs{P}$ and $0\leq j \leq \abs{T}$. We may compute the matrix $D$ with the DP scheme
\begin{align}
D(i, j) = \min 
\begin{cases}
\label{eqn:wf:dmin}
c(p_i \rightarrow \epsilon) + D(i-1, j) \\
c(p_i \rightarrow t_j) + D(i-1, j-1) \\
c(t_j \rightarrow \epsilon) + D(i, j-1)
\end{cases}
\end{align}
The $\min$ is over terms with valid indices into $D$ (\eg\ if $i>0$ and $j=0$, only the top term is considered) --- this notational convenience simplifies special cases $D(i,0)$ and $D(0,j)$ as in \eg\ \cite{sellers}. After initializing
\begin{align}
\label{eqn:wf:edinit}
D(0, 0) = 0,
\end{align}
$\Ocal(\abs{P}\abs{T})$ work yields $d(P, T) = D(\abs{P}, \abs{T})$.
\newcommand\sellersfigt{GATCGTCGATC}
\newcommand\sellersfigp{GATC}

\begin{figure}%
  \begin{center}
  {
  \footnotesize
  \begin{tabular}{lrrrrrrrrrrrr}
\toprule
{} &  - &  G &  A &  T &  C &  G &  T &  C &  G &  A &  T &  C \\
\midrule
- &  0 &  0 &  0 &  0 &  0 &  0 &  0 &  0 &  0 &  0 &  0 &  0 \\
G &  1 &  0 &  1 &  1 &  1 &  0 &  1 &  1 &  0 &  1 &  1 &  1 \\
A &  2 &  1 &  0 &  1 &  2 &  1 &  1 &  2 &  1 &  0 &  1 &  2 \\
T &  3 &  2 &  1 &  0 &  1 &  2 &  1 &  2 &  2 &  1 &  0 &  1 \\
C &  4 &  3 &  2 &  1 &  0 &  1 &  2 &  1 &  2 &  2 &  1 &  0 \\
\bottomrule
\end{tabular}

  }
  \end{center}
  \caption{
    \label{fig:sellers}
    An example of the matching algorithm of \autoref{sec:ss:seller} with $T=\text{``\sellersfigt''}$ and $P=\text{``\sellersfigp''}$, and unit costs $c$. We tabulate $D(i,j)$ with the elements of $P$ (respectively $T$) as the row (column) labels. The dashes are place holders for the zero index labels. Note that the bottom row gives the edit distance from $P$ to the corresponding suffix of $T$ which ends at the given column, so that the two zeros therein correspond to exact matches of $P$.
  }
\end{figure}

\section{A Generic Autoregressive Sequence Model}
\label{sec:ss}

We now set up our computational architecture for the simpler analogous case of traditional edit distance, in two steps. In \autoref{sec:ss:seller} and \autoref{sec:ss:walder} we derive a DP for computing the edit distance from the suffix to all other sub-strings. Then in \autoref{sec:ss:matching} we present a general scheme for utilizing the distances computed in the previous step, for forecasting (and therefore sequence modelling).

\subsection{Dynamic Program for String Matching}
\label{sec:ss:seller}

Sellers' modification \citep{sellers} of the DP of \autoref{sec:ed:wf} involves initializing $D(0, j)=0$ for all $j$ (rather than just $j=0$ as per \eqref{eqn:wf:edinit}), and applying  \eqref{eqn:wf:dmin} for $i > 0$; see \eg\ \autoref{fig:sellers}. The bottom row $D(\abs{P}, j)$ gives the minimum edit distance from $P$ to $T(j':j)$ for any $j'\leq j$. That is, we match only the suffix $T(j':j)$ of $T(:j)$.

\newcommand{\assign}{\,\,\leftarrow \,\,}
\begin{algorithm}[t]
   \caption{MotifNet generalised distance.}
   \label{alg:distance}
\begin{algorithmic}
   \STATE {\bfseries Input:} 
   $S=s_1 \cdot s_2 \cdots s_{\abs{S}},\fembedding,\faddition,\fsubstitution,\fdeletion,\dinit$
   \STATE {\bfseries Output:} $D(i, j, k)$ 
   \FOR{$i=1$ {\bfseries to} $\abs{S}$}
   \FOR{$k=1$ {\bfseries to} $i$}
   \IF{$k = 1$}
   \FOR{$j=1$ {\bfseries to} $i$}
   \STATE $D(i,j,k)\assign \faddition(\dinit,\, \fsubstitution(s_i, s_j))$
   \ENDFOR
   \ELSE
   \FOR{$j=1$ {\bfseries to} $i$}
   \STATE $D_\downarrow \hspace{1.39mm} \assign \faddition(D(i-1,j,k-1),\, \fdeletion(s_i))$
   \STATE $D_\searrow \hspace{0mm} \assign \faddition(D(i-1,j-1,k-1),\, \fsubstitution(s_i, s_j))$
   \STATE $D_\rightarrow \hspace{0mm} \assign \faddition(D(i,j-1,k),\, \fdeletion(s_j))$
   \vspace{2mm}
  \STATE $D(i,j,k) \assign {
  {\displaystyle \argmax_{D'\in \left\{ D_\downarrow, D_\searrow, D_\rightarrow \right\}}} \fscoring(D')}$%
   \ENDFOR
   \ENDIF
   \ENDFOR
   \ENDFOR
\end{algorithmic}
\end{algorithm}

\begin{figure}
\begin{center}
\includegraphics[height=0.33\textwidth]{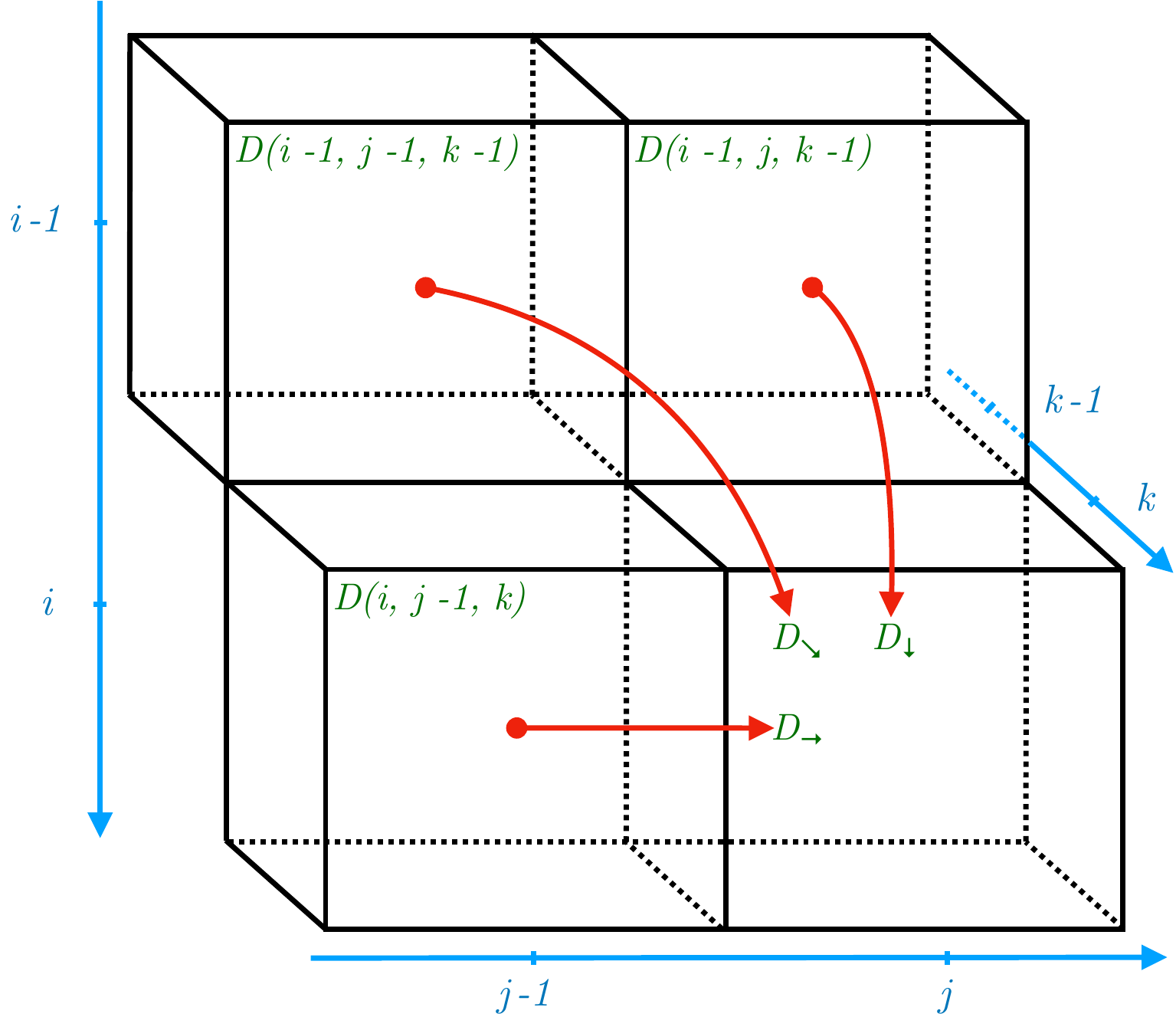}
\end{center}
\caption{
\label{fig:blocks} 
A diagram of the information flow (red arrows) in the generalised distance tensor $D(i,j,k)$, as per \eqref{eqn:ss:walder:dmin} and algorithm \ref{alg:distance}. The red arrows correspond to edit operations; \eg\ deletion of $s_j$ in the case of the lower red arrow.
}
\end{figure}

\subsection{Self Matching}
\label{sec:ss:walder}

Given a sequence $S$, let, for $j\leq i$, $\ds(i,j,k)$ be the minimum edit distance from $S(i-k+1:i)$ (roughly, the subsequence of length $k$ ending at $i$; recall \autoref{sec:notation}) to the suffix $S(j':j)$ of $S(:j)$, for any $j'\leq j$. We interpret $\ds(i,j,k)$ as the shortest distance from the subsequence of length $k$ ending at $i$ to the suffix ending at any $j$ satisfying
\begin{align}
\label{eqn:jlti}
j\leq i.
\end{align}

To derive a recursion analogous to \eqref{eqn:wf:dmin}, we apply the scheme of \autoref{sec:ss:seller}. Let $D_{X,Y}$ be the matrix obtained by applying the scheme of \autoref{sec:ss:seller} with $P=X$ and $T=Y$. Furthermore let $P_i=S(i-k+1:i)$ and $T_j=S(:j)$. This notation allows the precise definition
\begin{align}
\ds(i,j,k) & = D_{P_i, T_j}(\abs{P_i}, \abs{T_j}),
\end{align}
as well as the relations
\newcommand{\spc}{=}
\newcommand{\spcb}{}
\begin{align}
\spcb \ds(i-1,j,k-1) & \spc D_{P_i, T_j}(\abs{P_i}-1, \abs{T_j})\\
\spcb \ds(i,j-1,k) & \spc D_{P_i, T_{j-1}}(\abs{P_i}, \abs{T_{j-1}}).
\end{align}

Combining these relations with \eqref{eqn:wf:dmin} we obtain that, for $k>1$
\begin{align}
\label{eqn:ss:walder:dmin}
\ds(i, j, k) = \min 
\begin{cases}
c(s_i \rightarrow \epsilon) + \ds(i-1, j, k-1) \\
c(s_i \rightarrow s_j) + \ds(i-1, j-1, k-1) \\
c(s_j \rightarrow \epsilon) + \ds(i, j-1, k).
\end{cases}
\end{align}
In this case the $\min$ is over terms for which both
\begin{enumerate}
\item the indices into $\ds$ are non-negative (as in \eqref{eqn:wf:dmin}), and
\item the second index into $\ds$ is less than or equal to the first. This affects the first term only, and for $i=j$ only, and is due to the assumption \eqref{eqn:jlti}. Without this condition, slices $\ds(i,:,k)$ would depend on $s_{i+1}$, violating temporal causality in the model which follows. 
\end{enumerate}
One may view $s_i\rightarrow\epsilon$ (resp. $c(s_j\rightarrow\epsilon)$) as insertion (resp. deletion), or vice versa, depending on the perspective. We model them symmetrically and refer to both as \textit{deletion}. 
\subsection{Forecasting / Sequence Modelling}
\label{sec:ss:matching}

Since in general 
\begin{align}
\label{eqn:chainrule}
p(S)=\smash[t]{\prod_{i=0}^{\abs{S}-1}} p(s_{i+1}|S(:i)),
\end{align}
we may model a sequence $S$ by assuming $S(:i)$ is given, and predicting the next element $s_{i+1}$. A natural approach is to compare suffixes $S(i-k+1:i)$ for various $k$, to the previous sub-sequences $S(j':j)$ for some $j'\leq j \leq i$. 
Hence $D(i, j, k)$ is useful: an exact match $\ds(i,j,k)=0$ for some $j$ and $k$, may suggest the continuation $s_{j+1}$ to re-occur (that is,  equal $s_{i+1}$). Generally, we may forecast using all $D(i, j, k)$ and corresponding next symbols $s_{j+1}$, for $1 \leq j < i$ and $1 < k < i$ by modelling 
\begin{align}
\label{eqn:ss:forecast}
s_{i+1} \,|\, S(:i) \sim 
\mathcal{S}
\lnb
\lcb 
(D(i,j,k), s_{j+1})
\rcb_{0 \leq j < i, \, 0 < k < i}
\rnb \,\,\,\,\,\,
\end{align}
where $\mathcal{S}$ is a random variable parametrised by a set of (distance, symbol) pairs; \eg\ $\mathcal{S}$ may place higher probability on symbols $s_{j+1}$, with smaller corresponding $D(i,j,k)$.

This scheme subsumes various sequence prediction algorithms, \eg\ variable length Markov models, prediction suffix trees, and  on-line sequence prediction algorithms \cite{Willems95thecontext,jordanmmmm,dekel}.

\section{Motif Networks}
\label{sec:model}

\begin{table}
\center
\subfigure[functions]{
\begin{tabular}{llll}
\toprule
Notation & Role & Architecture & Mapping \\
\midrule
\fembedding & embedding & lookup & $\Sigma \rightarrow \mathcal{E}$ \\
\fdeletion & deletion & FF & $\Sigma \rightarrow \mathcal{C}$ \\
\fsubstitution & substitution & FF & $\Sigma \times \Sigma \rightarrow \mathcal{C}$ \\
\faddition & addition & GRU & $\mathcal{D} \times \mathcal{C} \rightarrow \mathcal{D}$ \\
\fscoring & scoring$^\star$ & FF & $\mathcal{D} \rightarrow \mathbb{R}$ \\
\fanalogy & analogy & FF & $\mathcal{D} \times \mathcal{E} \rightarrow \mathcal{O}$ \\
\fforecasting & forecasting & FF & $\mathcal{O} \rightarrow \mathbb{R}^{\abs{\Sigma}}$ \\
\bottomrule
\end{tabular}
}
\subfigure[sets \& spaces]{
\begin{tabular}{ll}
\toprule
Notation & Interpretation of Elements \\
\midrule
$\Sigma=\lcb 1, 2, \dots , \abs{\Sigma}\rcb$ & Discrete symbol \\
$\mathcal{E}=\realset^{N_E}$ & Embedding \\
$\mathcal{C}=\realset^{N_C}$ & Generalised edit cost \\
$\mathcal{D}=\realset^{N_D}$ & Generalised distance \\
$\mathcal{O}=\realset^{N_O}$ & Penultimate layer \\
\bottomrule
\end{tabular}
}
\caption{
\label{table:functions} 
Our function modules and the spaces they act on. \textit{FF} and \textit{GRU} stand for \textit{feed-forward} and \textit{gated recurrent unit}, respectively.  $^\star$Note that the scoring function $\fscoring$ performs several roles.
}
\end{table}

The previous \autoref{sec:ss} presented a general framework for sequence prediction. We now generalise that framework to obtain our novel deep learning architecture. The  components of our model are summarized in \autoref{table:functions}.

\subsection{Motif Network Distance Tensor}
\label{sec:model:distance}

The neural dynamic program for $D(i,j,k)$ is given as algorithm \ref{alg:distance}, and utilises the following learned modules:
\begin{itemize}
    \item  \textbf{Embedding vectors} $\lcb \bm{e}_s\rcb_{s\in\Sigma}\subset \mathcal{E}$ of our symbols $s\in\Sigma$, so that $\fembedding(s)=\bm{e}_{s}$.
    \item The \textbf{cost of edit operations} is learned using  feed-forward neural networks. For \textbf{deletion} $s_i\rightarrow \epsilon$ we let, in the notation of \autoref{sec:notation}, $\fdeletion=\gdeletion \circ \fembedding$ where $\gdeletion=\relu_\alpha \circ \, L(A_D^{(2)}) \circ \relu_\alpha \circ \, L(A_D^{(1)})$. The \textbf{substitution} operation $s_i\rightarrow s_j$ is slightly more subtle; for reasons of efficiency (see \autoref{sec:complexity}) and parsimony, we employ a \textit{symmetric} function of two arguments. Symmetry is obtained via a differentiable approximation to the absolute value function, namely $\pseudohuber$ of \autoref{sec:notation}. In particular we let $\fsubstitution(s_i,s_j)=\gsubstitution(\pseudohuber(\fembedding(s_i)-\fembedding(s_j)), \delta))$, where $\gsubstitution$ is of the same form as (but does not share parameters with) $\gdeletion$, and $\delta=\frac{1}{2}$ throughout. 
    \item We let the elements of the \textbf{distance tensor} $D(i, j, k)$ be vector rather than scalar, and we denote this space of generalised distances by $\mathcal{D}$. This requires  generalization of the notion of \textbf{addition} on the r.h.s. of \eqref{eqn:ss:walder:dmin}; we define $\faddition:\mathcal{D}\times\mathcal{C}\rightarrow\mathcal{D}$, and we assume a Gated Recurrent Unit architecture (GRU) \cite{gru} where  distance is the latent state, with initial value $\dinit$. The GRU is slightly more appropriate than the LSTM cell in that we don't require an output gate. 
    \item We generalise the $\min$ in \eqref{eqn:ss:walder:dmin} to $\arg\max_D \fscoring(D)$ where $\fscoring=L(A_W)$ is a scalar valued  \textbf{scoring} function (\ie\ $A_W\in\realset^{N_D\times 1}$). A soft max alternative is addressed in the ablative study of \autoref{sec:experiments:toy}.
\end{itemize}

\subsection{Motif Network Forecast}
\label{sec:model:forecast}

We generalise \eqref{eqn:ss:forecast} in a manner designed to model the various self similarities inherent in symbolic music. The key concept we introduce is that of the \textit{analogy} function $\fanalogy(\bm{d},\bm{e})=\ganalogy((\bm{d}\ttran, \bm{e}\ttran)\ttran)$. Here $\ganalogy$, which acts on the concatenation of $\bm{d}$ and $\bm{e}$, is a two layer feed-forward network, $\ganalogy=\relu_\alpha \circ \, L(A_A^{(2)}) \circ \relu_\alpha \circ \, L(A_A^{(1)})$. 

We apply the analogy function in this way to all motif positions $j$ and lengths $k$, and take a weighted average, so 
\begin{align}
\label{eqn:oi}
O_i=\sum_{0\leq j < i} \sum_{0 \leq k < i} w_{i,j,k} \fanalogy(D(i,j,k), \fembedding(s_{j+1})),
\end{align}
where the weights are obtained by applying a softmax to $\fscoring(D(i,j,k))$, with the same $\fscoring$ as algorithm \ref{alg:distance}:
\begin{align}
\label{eqn:wijk}
w_{i,j,k}= \frac{\exp(\fscoring(D(i,j,k)))}{\sum_{0\leq j' < i} \sum_{0 \leq k' < i} \exp(\fscoring(D(i,j',k')))}
\end{align}
The conditional probability mass function for $s_{i+1}|S(:i)$ on the r.h.s. of \eqref{eqn:chainrule} is then given by $\fforecasting(O_i)$, where the forecasting function is another two layer feed-forward neural network, $\fforecasting=\sigma \circ \, L(A_F^{(2)}) \circ \relu_\alpha \circ \, L(A_F^{(1)})$. 

\subsubsection{Analogy function \fanalogy : intuition} 
\label{sec:model:forecast:analogy}

Given \eg\ an earlier sub-sequence $3,7,5,7$, what likely follows $13,17,15$? Here, our generalised distance should encode the transformation ``add  10'', which \fanalogy\ should apply to $7$ yielding 17. In line with \eqref{eqn:ss:forecast}, $\fanalogy$ acts on $D(i, j, k)$ and $\fembedding(s_{j+1})$, and returns information pertinent to predicting $s_{i+1}$. For another example, with $i$ and $j$ as marked in \autoref{fig:bach}, and suffix length $k=6$, $D(i,j,k)$ captures the relationship (\textit{generalised distance} is a slight misnomer) between the first six and last six notes. Given the earlier continuation $s_{j+1}$, the analogy function may predict that $s_{i+1}$ would complete the analogy (in this case we expect the second last note $s_{i-1}$ to re-occur as $s_{i+1}$).

\subsection{Combining with a Recurrent Neural Network}
\label{sec:model:combination}

We may combine the Motif Network with a traditional recurrent neural network such as that of the Long Short Term Memory (LSTM) \cite{lstm}. To this end we apply the usual LSTM recurrence with the embeddings $\fembedding(s_i)$ as the input for the $i$-th time step. We then concatenate the LSTM output with the Motif Network output $O_i$ of \eqref{eqn:oi} before applying $\fforecasting$.

\newcommand\aaa{\textit{a)}}
\newcommand\aaaa{\textit{(a)}}
\newcommand\bbb{\textit{b)}}
\newcommand\bbbb{\textit{(b)}}
\begin{figure}[t]
\begin{center}
    \includegraphics[height=0.42\textwidth,page=9]{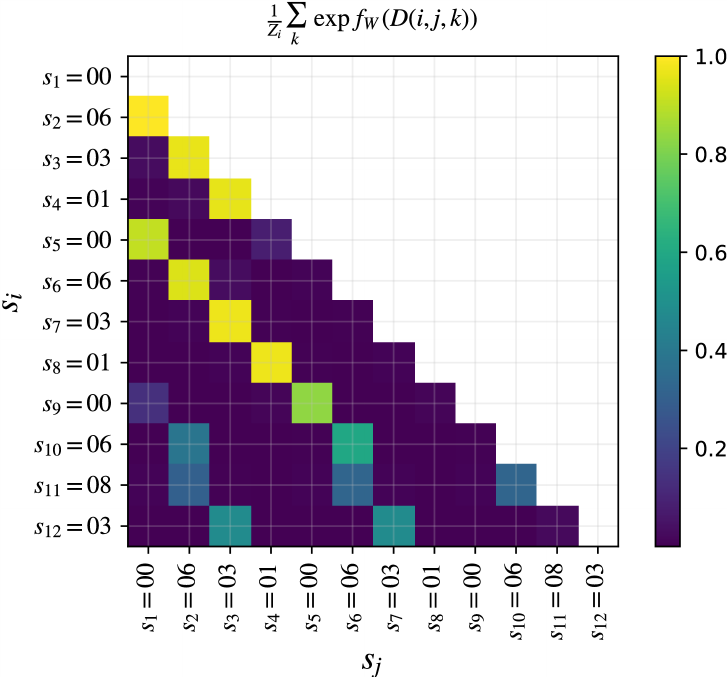}
\end{center}
\caption{
\label{fig:weight} 
Visualising the scoring function \fscore\ on a test sequence from the \texttt{markov\_\,editloop} toy dataset. Brightness indicates alignment for all rows $i$ and columns $j<i$, summed over length $k$ and normalised (see the title). Values $s_i$ and $s_j$ are given in the axis labels. 
Rows $i=5$ to $i=8$ align the motif, while rows $i>8$ average over two valid alignments. 
The insertion noise $s_{11}=8$ is handled as evidenced by the similarity between rows $i=10$ and $i=11$. See \autoref{sec:experiments:toy} for more details.
}
\end{figure}

\section{Efficient Approximation with Trees}
\label{sec:trees}

From \eg\ \eqref{eqn:oi} we see that the cost of forecasting $s_{i+1}$ given $S(:i)$ is quadratic in $i$, leading to a cubic time complexity $O(\abs{S}^3)$ for the entire sequence. These calculations involve redundancies which we can exploit. In contrast to existing literature focusing on suffix trees \cite{Ukkonen93approximatestring,Ukkonen1995,Willems95thecontext,dekel}, the natural data structure in our case is the \textit{edit} tree. This is a natural consequence of modelling transformations between arbitrary sub-sequences of a sequence, rather than the relationship between immediately subsequent sub-sequences.

\subsection{Edit Trees}

From algorithm \ref{alg:distance} it is evident that each $D(i,j,k)\in\mathcal{D}$ is a function of a sequence of edit operations which correspond to an alignment of sub-strings of $S$. We track these edit operations using a rooted tree $G=(V,E)$ whose edges $E= \Sigma \times\lnb\Sigma\cup\epsilon\rnb$ correspond either to deletion $s\rightarrow \epsilon;\,\, s\in\Sigma$ or substitution $s\rightarrow s';\,\, s,s'\in \Sigma$. Let $\nu:V\rightarrow\mathcal{D}$ map nodes to distances, and (loosely speaking) let $\fedit(e)$ take on the value $\fsubstitution(e)$ (respectively  $\fdeletion(e)$) for $e$ corresponding to substitution (respectively deletion). For all vertices $v,v'$ connected by $e$ and satisfying $\parent(v')=v$, $\nu$ satisfies $v(\roooot(G))=D_0$ as well as the recursion $\nu(v')=\faddition(\nu(v),\fedit(e));\,\,v'{}\neq\roooot(G)$.  

\subsection{Priority Queue}
\label{sec:priority}

We assume that the number of useful alignments is small compared to the quadratic total number. These fruitless alignments are benign in that they may obtain insignificant weights $w_{i,j,k}$, as learned by the function $\fscoring$, and thereby have little effect on the forecast due to \eqref{eqn:wijk}.

This suggests the following approximation. Rather than computing the entire tree, we expand only the most promising branches using $\fscoring$ as our heuristic. The mechanism we propose is to only expand children from those nodes which have a heuristic score which is among the \npriority\ best among its siblings. More precisely, when algorithm \ref{alg:distance} would lead to a new vertex $v$ being created in $G$ as a child of some vertex $v'$ (so that $v'=\parent(v)$), we compute the descending rank order statistic of the associated heuristic $\fscoring(\nu (v))$, among that of the extant siblings of the parent, $\lcb \fscoring(v'')\rcb_{v''\in \siblings(v')}$. If the rank exceeds some parameter \npriority, then $v$ is omitted from the tree, and the associated elements of $D(i,j,k)$ (those satisfying $D(i,j,k)=\nu(v)$) are omitted from \eqref{eqn:oi} and \eqref{eqn:wijk}. This is the third role assumed by $\fscoring$ (after \eqref{eqn:wijk}, and the $\arg \min$ in algorithm \ref{alg:distance}).

\subsection{Bounded Tree Depth and Suffix Length}
\label{sec:boundeddepth}

We further assume that the length of alignment need not exceed a threshold, \dmax. Vertices beyond this tree depth are omitted, along with associated terms in \eqref{eqn:oi} and \eqref{eqn:wijk}. We further bound by \dmax\ the length of suffix used for prediction, thereby considering only $D(i,j,k)$ for $k \leq \dmax$.


\begin{figure*}[t!]%
  \begin{center}
  {
  \includegraphics[width=0.86\textwidth]{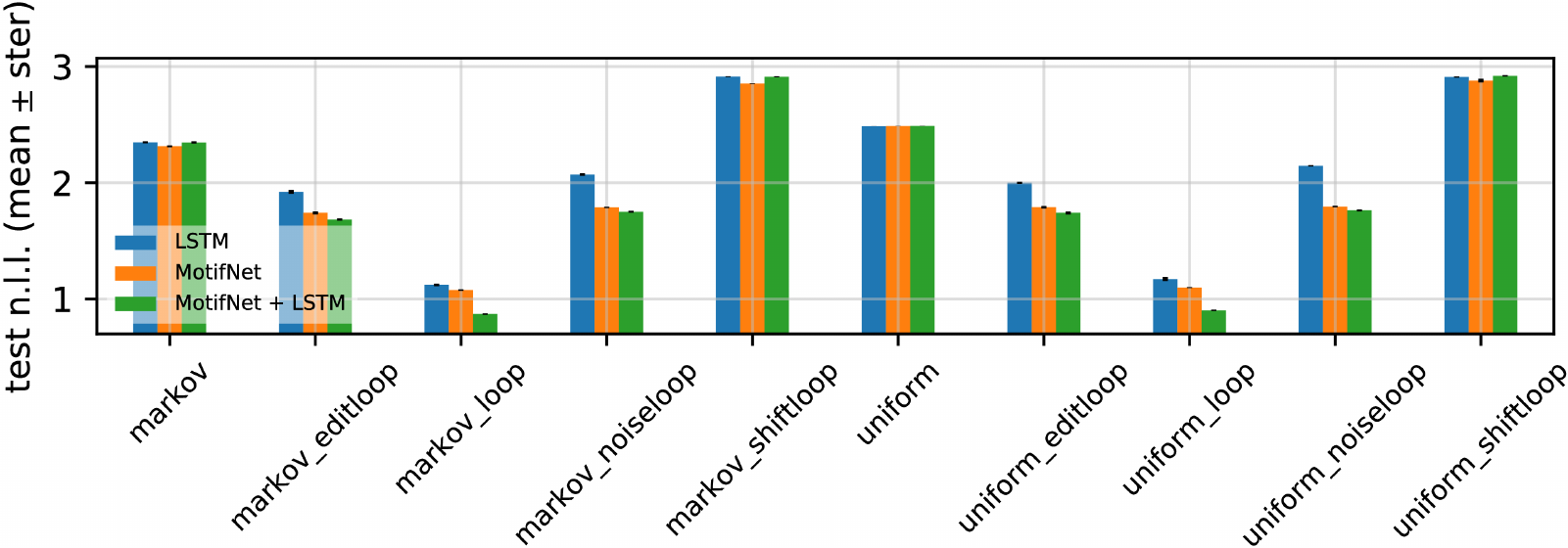}
  }
  \end{center}
  \caption{
  Average test set negative log likelihood (with barely visible $\pm 1$ \textit{standard error} bars) for a stacked LSTM, MotifNet, and their combination (see \autoref{sec:model:combination}), on a suite of toy problems. See \autoref{sec:experiments:toy} for more details.
    \label{fig:toy_table} }
\end{figure*}

\subsection{Computational Complexity}
\label{sec:complexity}
The maximum fan-out of the edit tree is $O(\abs{\Sigma}^2)$ due to there being $\abs{\Sigma}(\abs{\Sigma}-1)/2$ unique substitutions (recall that $\fsubstitution$ is symmetric), and $\abs{\Sigma}$ deletions. By limiting the tree depth the required number of tree nodes (and associated unique generalised distance tensors) is therefore $O(\abs{\Sigma}^{2\dmax})$. The forecast \eqref{eqn:oi} combines these distance tensors with the other argument $\fembedding(s_{j+1})$ (of which there are $\abs{\Sigma}$ unique values) to the analogy function $\fanalogy$. This is done per time-step, leading to an overall time complexity of $O(\abs{S} \abs{\Sigma}^{2\dmax+1})$. 

In the worst case, the priority queue does not reduce the effective fan-out of the tree (as the children may be added in ascending order of the value of the heuristic $\fscoring$). In the best case, the number of required nodes is $O((\npriority)^{\dmax})$. Including a factor of $\abs{\Sigma}$ due to \eqref{eqn:oi}, we get an overall best case complexity of $O(\abs{S} \abs{\Sigma} (\npriority)^{\dmax})$. The real runtime is therefore heavily affected by the actual data distribution --- see \autoref{sec:experiments:music} for an empirical investigation.


\begin{figure*}[t]%
  \begin{center}
  {
  \includegraphics[width=0.99\textwidth]{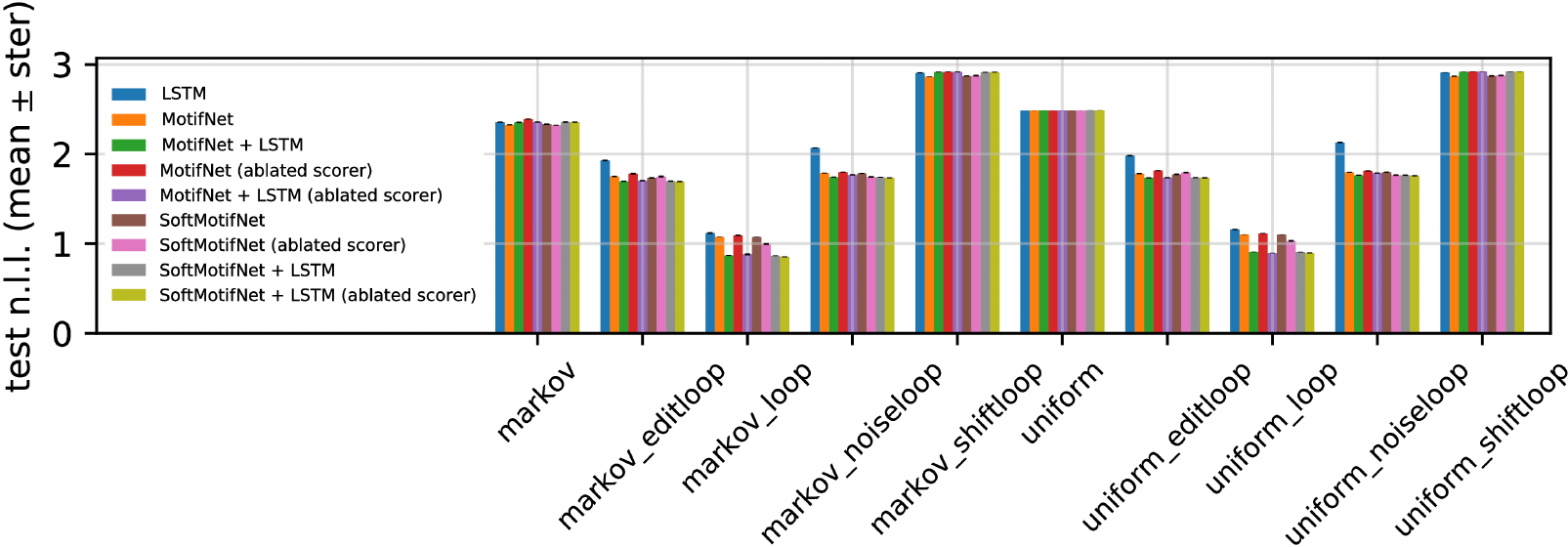}
  }
  \end{center}
  \caption{
  An expanded \autoref{fig:toy_table} which includes variations of our MotifNet. See \autoref{sec:experiments:toy} for more details.
    \label{fig:toy_table_expanded}  }
\end{figure*}

\section{Experiments}
\label{sec:experiments}

\subparagraph{Methodology.} We used a train/validation/test scheme based on log likelihoods. We train full epochs up to three  \textit{strikes} (increases in validation set log likelihood after a training epoch). Training and validation was performed for a range of hyper-parameters. We let the spaces $\mathcal{E}, \mathcal{C}, \mathcal{D}$ have the same dimension, which we varied as $2, 4, 8, \dots, 2048$. Three algorithms are compared throughout: the \textbf{LSTM} \cite{lstm}, our \textbf{MotifNet}, and the combination \textbf{MotifNet+LSTM} of \autoref{sec:model:combination}.

\subparagraph{Implementation.} Our implementation\footnote{\texttt{https://bitbucket.org/cwalder/motif}} relied heavily on the dynamic graph of the \textit{PyTorch} software; nonetheless we found the tree based implementation of MotifNet to be rather more involved than, say, the LSTM. We trained with stochastic gradient descent using  \textit{Adam}  \cite{adamoptimiser}. GPUs yielded only modest speed ups, so we worked with a CPU cluster. Proper parallelization of MotifNet is the subject of ongoing research, and our results suggest that the full power of MotifNet remains to be revealed (see \autoref{fig:music_results_figure}).

\subparagraph{Key parameters.} We allowed the LSTM variants an advantage by letting the LSTM cell be \textit{stacked} \cite{stackedlstm,stackedlstmb} with number of layers ranging 1, 2, 3, 4, whereas for the MotifNet we fixed $\faddition$ to be a GRU with one layer. For the toy problems of \autoref{sec:experiments:toy} we used an exact MotifNet; for the music data of \autoref{sec:experiments:music} we used an approximate MotifNet with the crude setting of $\dmax=4$, and \npriority\ ranging $2, 4, 8, \dots, 256$. The hyper-parameters with best validation likelihood (per data replicate in the case of the toy data) were applied to the test set. We further ensured that the best models occurred well within the range considered for each parameter above, with the exception of the \npriority\ parameter (for which MotifNet may clearly perform even better, with larger but more computationally expensive settings --- see \autoref{fig:music_results_figure}).

\subsection{Toy Problems}
\label{sec:experiments:toy}

\subparagraph{Data.} We considered two toy processes. \textbf{\texttt{uniform}} generates each symbol $s_i$ independently from the uniform distribution on the base alphabet $\Sigma_0=(0,1,\dots,11)$. While \texttt{uniform} has no structure, we generate structured sequences by repeating motifs generated by it. \textbf{\texttt{markov}}, is a Markov chain with initial and transition probabilities drawn uniformly at random. For each process and the five generation schemes below, we generate 300 sequences each for training, validation and testing. The process is repeated for each of 16 replicates (each of which with different \texttt{markov} parameters). The five generation schemes were:
\begin{itemize} 
    \item \textbf{(no label)}: draw a sequence of length $\abs{S}=12$.
    \item \textbf{\texttt{loop}}: a motif of length 4 is drawn from the process and repeated three times. For example $S=(0,1,5,10,0,1,5,10,0,1,5,10)$.
    \item \textbf{\texttt{shiftloop}}: similar to \texttt{loop} but each repetition of the base motif is shifted by an integer drawn uniformly from (0,1,\dots,11). For example $S=(0,1,2,10,0+2,1+2,2+2,10+2,0+1,\etc)$.
    \item \textbf{\texttt{noiseloop}}: Similar to \texttt{loop}, but each element of the sequence is randomly (with probability 0.15) replaced with a uniform sample from $\Sigma_0$.
    \item \textbf{\texttt{editloop}}: Similar to \texttt{loop}, but for each element $s_i$ we randomly (with probability 0.15) either (with probability 0.5) delete $s_i$ or (with probability 0.5) insert a new element (drawn uniformly from $\Sigma_0$) after $s_i$.
\end{itemize}

\subparagraph{Visualisation.} In \autoref{fig:weight} we demonstrate that the learned $\fscoring$ aligns the sequence with itself as intended.

\subparagraph{Numerical results.} we compare test scores in \autoref{fig:toy_table}
. As intended MotifNet better captures the regularity of loops in all cases. The combination MotifNet+LSTM adds a further small improvement, with the exception of \texttt{shift\_loop}, where the out-performance of MotifNet is already very slight (disappointingly; this requires investigation).

\subparagraph{Ablative Study.} In \autoref{fig:toy_table_expanded} 
we consider the following MotifNet variations. The first, \textit{ablated scorer}, learns a separate \fscorer\ for the  recursion \eqref{eqn:wf:dmin} and forecasting (\ref{eqn:oi}--\ref{eqn:wijk}). The second, \textit{Soft}, replaces the hard max of \eqref{eqn:wf:dmin} with a softmax based on \fscorer\ (similarly to (\ref{eqn:oi}--\ref{eqn:wijk}), for example). The results demonstrate the point of coupling \fscorer\ across these two distinct roles in MotifNet (recall there is even a third role of \fscorer, namely tree pruning as per \autoref{sec:trees}). Indeed, we find that decoupling breaks the algorithm (intuitively, the hard max \eqref{eqn:ss:walder:dmin} does not permit gradient flow). While introducing the softmax heals this breakage, it also leads to non sparse gradients and, fatally, makes the tree approximation of \autoref{sec:trees} impossible --- the reason for this being that under the softmax, elements $D(i,j,k)$ would depend on all of the possible transitions (red arrows in \autoref{fig:blocks}). Importantly, the softmax with ablative scorer does not outperform (interestingly, it slightly under-performs) the basic (and computationally tractable) scheme --- \eg\ compare \textit{MotifNet} and \textit{SoftMotifNet (ablated scorer)} in \autoref{fig:toy_table_expanded}.

\subsection{Symbolic Music Data}
\label{sec:experiments:music}

\newcommand{\pfig}{2018_02_05_15_21_13_743520_948646797}
\newcommand{\ww}{0.75}

\begin{figure*}%
  \begin{center}
  \subfigure[test set negative log likelihood]{
  \includegraphics[page=2,width=\ww\textwidth]{figs/raw/Css/\pfig/toy-crop-raster.pdf}
  }
  \subfigure[median time per epoch]{
  \includegraphics[page=3,width=\ww\textwidth]{figs/raw/Css/\pfig/toy-crop-raster.pdf}
  }  \end{center}
  \caption{
  Trading MotifNet (embedding dimensionality 2048) accuracy (a) for speed (b) with  \npriority\ of \autoref{sec:trees}.  See \autoref{sec:experiments:music}.
    \label{fig:music_results_figure}
  }
\end{figure*}


\newcommand{\astamp}{2018_02_05_12_01_12_828911_912871497}
\newcommand{\bstamp}{2018_02_05_12_04_27_147348_957862267}
\newcommand{\cstamp}{2018_02_05_12_05_19_855326_690152391}
\newcommand{\lstmtablename}{LSTM}
\newcommand{\motifnetlstmtablename}{MotifNet+LSTM}
\newcommand{\motifnettablename}{MotifNet}

\begin{table}%
  \caption{
  Average test set negative log likelihood for a stacked LSTM, MotifNet, and their combination (see \autoref{sec:model:combination}) on real symbolic music problems. See \autoref{sec:experiments:music} for more details.
    \label{fig:music_results_table}
  }
  \begin{center}
  \begin{tabular}{lllll}
\toprule
  & JBM & MUS & NOT & PMD \\ 

\midrule
\lstmtablename & 
1.82 & 2.03 & 1.03 & 2.67 \\

 \motifnettablename & 
1.77 & 1.88 & 0.81 & 1.90 \\

 \motifnetlstmtablename & 
1.79 & 1.83 & 0.73 & 1.85 \\

\bottomrule
\end{tabular}
  \end{center}
\end{table}

\subparagraph{Data.} We used the same four sets of midi files as \cite{bl}, but rather than deriving simplified piano rolls, we derived simplified note onset sequences. The Bach chorale midis of \cite{bl} lack valid channel data, so we downloaded the analogous files from \cite{musedata} for that dataset. The four datasets are \textbf{JBM} (J.S. Bach chorales from \cite{musedata}), \textbf{MUS} (the MuseData set of \citet{bl}), \textbf{NOT} (Nottingham chord data of \citet{nottinghamdata} converted to midi by \citet{bl}) and \textbf{PMD} (piano midis provided by \citet{pmddata}).

\subparagraph{Preprocessing.} We split the files by midi channel (or instrument).  We ordered the midi \textit{onset} events (discarding other event types) by increasing time and then pitch, and retained only the midi number in our final test sequence. A more sophisticated application of our MotifNet to music generation is the subject of non-trivial ongoing work.

\subparagraph{Numerical results.}  MotifNet out-performs the LSTM benchmark (see \autoref{fig:music_results_table}). Recall that in addition to tuning the embedding dimensionality, for LSTM we tune the number of layers, whereas for MotifNet we merely tune \npriority\ of \autoref{sec:trees}. This is significant; from \autoref{fig:music_results_figure} see that \npriority\ does indeed merely trade computation time for accuracy. Moreover while the performance has not yet plateaued up to a value of 256, we already out-perform the LSTM. We also simply fixed the \dmax\ hyper-parameter of \autoref{sec:trees}. The results we present for MotifNet are at the limit of what is computationally convenient. Further scaling up in terms of dataset size and computational effort (as parametrised by \eg\ \npriority ) requires  algorithmic and implementation advances which are the subject of ongoing research.

\subparagraph{Discussion.} Recall that $\dmax=5$ limits the motif alignment length to 5 edit operations. It is interesting that this rather small value is sufficient to beat the LSTM on all music datasets. We conjecture that the reason the performance gap is smallest on the JBM set is that this dataset mainly features rather short sequences with little self similarity. As such, it is interesting that MotifNet can even match the LSTM on the JBM set, and seems to suggest that in the absence of strong motif alignments exploit, some global structure (key, harmonic sequence, \etc ) is partially captured. 

\section{Summary}
\label{sec:conclusion}

MotifNet combines a generalised edit distance recursion (between a sequence and itself) with an \textit{analogy} based forecasting rule. This captures regularities in the relationships between non-adjacent sub-sequences within the same sequence. The model is qualitatively different to traditional recurrent neural networks which, notwithstanding the power of the LSTM architecture, tend to focus on the relationship between sub-sequences and their immediate continuation. 

While \naively\ cubic in time complexity, the computational dependencies of the MotifNet lend themselves to representation and approximation by an \textit{edit tree}. This is a novel data structure which has edit operations (insertion, deletion, \etc ) as edges. By partially expanding this tree using a learned heuristic function, MotifNet is able to effectively model real music sequence data. 

An important novelty is the reuse of the scoring function \fscorer , the hardest working function in our model. Learning of \fscorer\ is permitted by the gradients which back propagate from the loss function via \eqref{eqn:oi}. The learned \fscorer\ then further serves by dictating the alignments found by the dynamic program of algorithm \ref{alg:distance}). The function also affords computational tractability through the termination of unpromising alignments as per \autoref{sec:priority}.



\bibliographystyle{icml2018}
\bibliography{walder}

\end{document}